(Research Article)

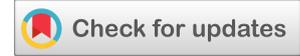

# Improving Large Language Model (LLM) fidelity through context-aware grounding: A systematic approach to reliability and veracity

Wrick Talukdar [1] and Anjanava Biswas [2, *]

[1] AWS AI and Machine Learning, Calgary, Alberta, Canada.
[2] AWS AI and Machine Learning, San Diego, California, USA.

**Abstract**

As Large Language Models (LLMs) become increasingly sophisticated and ubiquitous in natural language processing (NLP) applications, ensuring their robustness, trustworthiness, and alignment with human values has become a critical challenge. This paper presents a novel framework for contextual grounding in textual models, with a particular emphasis on the Context Representation stage. Our approach aims to enhance the reliability and ethical alignment of these models through a comprehensive, context-aware methodology. By explicitly capturing and representing relevant situational, cultural, and ethical contexts in a machine-readable format, we lay the foundation for anchoring a model's behavior within these contexts. Our approach leverages techniques from knowledge representation and reasoning, such as ontologies, semantic web technologies, and logic-based formalisms. We evaluate our framework on real-world textual datasets, demonstrating its effectiveness in improving model performance, fairness, and alignment with human expectations, while maintaining high accuracy. Furthermore, we discuss the other key components of the framework, including context-aware encoding, context-aware learning, interpretability and explainability, and continuous monitoring and adaptation. This research contributes to the growing body of work on responsible AI, offering a practical approach to developing more reliable, trustworthy, and ethically-aligned language models. Our findings have significant implications for the deployment of LLMs in sensitive domains such as healthcare, legal systems, and social services, where contextual understanding is paramount.

## 1. Introduction

LLM based Natural language processing (NLP) models have achieved significant advancements in recent years, enabling a diverse array of applications, including language translation, text summarization, dialogue systems, and content recommendation. However, the widespread adoption of these models has heightened concerns regarding their propensity to generate harmful or biased outputs, especially when deployed in sensitive contexts or high-stakes business domains. Contextual grounding, defined as the process of anchoring a model's understanding and decision-making within relevant situational, cultural, and ethical contexts, has emerged as a promising solution to these challenges. By explicitly incorporating contextual information, it is possible to better align a model's behavior with human expectations, societal norms, and ethical principles, thereby mitigating the risks of unintended or harmful consequences.

This paper proposes a novel framework for contextual grounding in textual models, aimed at enhancing their robustness, trustworthiness, and safety. Our approach focuses on the Context Representation stage, which involves developing methods to capture and represent relevant contextual information in a machine-readable format using knowledge representation techniques. We also provide an overview of the other key components of the framework, including context-aware encoding, context-aware learning, interpretability and explainability, and continuous monitoring and adaptation.

[*] Corresponding author: Anjanava Biswas, ORCID: https://orcid.org/0009-0008-3225-3316

## 2. Related Work

Contextual grounding has been explored in various domains within natural language processing (NLP), including sentiment analysis, dialog systems, and ethical language generation. Researchers have investigated how contextual factors like domain knowledge [1], cultural nuances [2], and situational contexts [3] impact the accuracy and fairness of models. Yushi Yao, Guangjian Li [4] proposed a context-aware sentiment analysis model that enhances sentiment word identification by incorporating contextual information. For dialog systems, studies have shown that incorporating contextual information such as conversational history, user profiles, and situational awareness can significantly improve the quality and naturalness of generated responses [5]. Zhaojiang et al. [6] introduced a context-aware neural conversation model that considers multi-turn dialog history and personal traits of the participants. In ethical language generation, the importance of contextual grounding has been recognized to ensure AI systems generate language aligning with societal norms, cultural values, and ethical principles [7].

Researchers at the Allen Institute for AI [8] proposed a framework for ethical language generation that incorporates social, cultural context, and ethical reasoning principles to avoid generating harmful or biased content. Additionally, studies emphasize the need to consider ethical and societal implications when developing and deploying AI systems [9, 10]. For instance, the AI Now Institute [10] highlights the importance of considering the broader societal impact, ethical implications, and potential risks associated with AI systems, particularly in sensitive domains or high-stakes applications. However, a comprehensive framework for contextual grounding in textual models, specifically focused on enhancing their robustness, trustworthiness, and safety, remains an area requiring further exploration and development. While previous works have addressed contextual grounding in specific domains or tasks, a unified approach integrating techniques from knowledge representation, adversarial training, interpretable AI, and continuous monitoring is lacking. Existing methods often focus on individual aspects of contextual grounding, such as incorporating domain knowledge or mitigating specific biases.

A holistic framework considering the interplay between situational, cultural, and ethical contexts, and their impact on model behavior across various tasks and applications, is needed to ensure the overall robustness, trustworthiness, and safety of textual models. By addressing this gap, the proposed framework in this paper aims to provide a comprehensive and principled approach to contextual grounding, enabling the development of more reliable, ethical, and trustworthy NLP systems that can be deployed in a wide range of real-world scenarios while aligning with human values and societal expectations.

## 3. Proposed framework

The proposed framework as displayed in Figure 1 below, comprises of five key components that work together to enable contextual grounding in textual models. Firstly, the Context Representation component captures and represents relevant contextual information, such as situational factors, cultural norms, and ethical considerations, in a machine-readable format using techniques from knowledge representation and reasoning. The Context-Aware Encoding component then integrates this contextual information with the textual inputs, ensuring that the model's understanding and decision-making are influenced by the relevant context. The Context-Aware Learning component incorporates contextual grounding into the model's training process, enabling it to learn context-sensitive representations and decision boundaries through techniques like multitask learning, adversarial training, and curriculum learning. To enhance transparency and scrutiny of the model's decision-making processes, the Interpretability and Explainability component leverages techniques from interpretable AI, such as attention visualization and counterfactual explanations. Finally, the Continuous Monitoring and Adaptation component implements mechanisms for continuous monitoring and adaptation of the model's behavior based on feedback and evolving contextual factors, using techniques from online learning and continual learning.

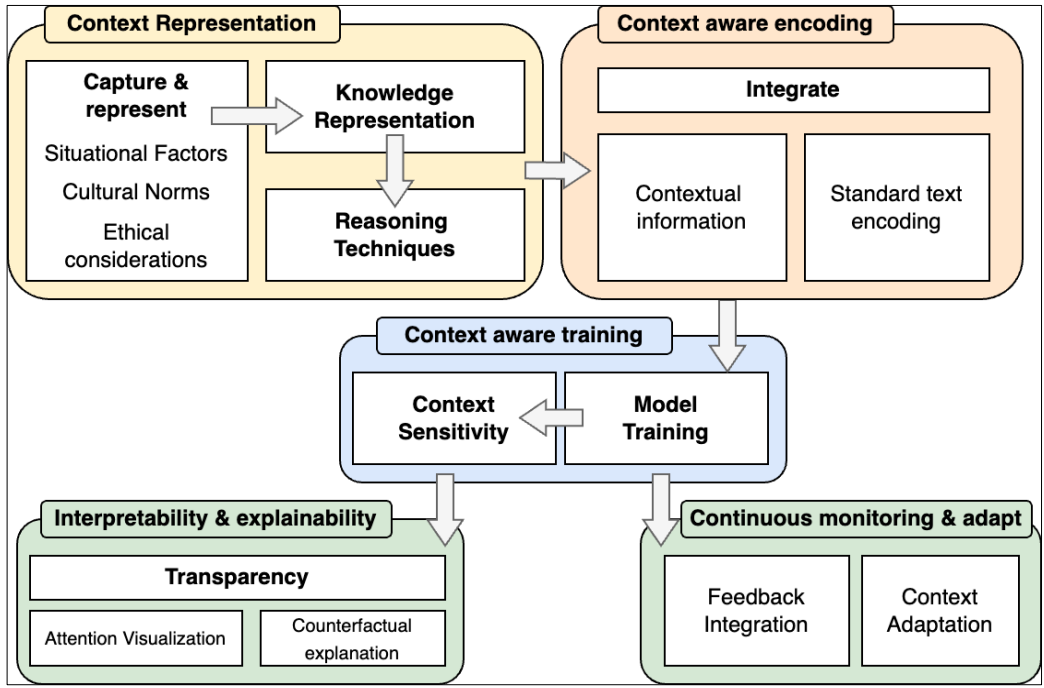

**Figure 1** Proposed framework

The Context Representation stage is the primary focus of this paper. We develop methods to capture and represent relevant contextual information, such as situational factors, cultural norms, and ethical considerations, in a machine-readable format. This involves leveraging techniques from knowledge representation and reasoning, including ontologies, semantic web technologies, and logic-based formalisms.

$$P(C) = \{ location = US, religion = Christian, age_{group} = Youth\} \ldots\ldots\ldots (1)$$

Let $C = \{c_1, c_2, \ldots, c_n\}$ be the set of context elements. Each context element $c_i \in C$ is represented as a set of predicates $P(c_i)$ and their associated values.

### 3.1. Situational Context

We developed ontologies to represent situational factors, such as location, time, activity, and environmental conditions. These ontologies capture the relationships and constraints among various situational elements, enabling context-aware reasoning and inference. Example ontology for representing situational contexts can be represented with formulas (2), (3), (4), (5) and (6).

$$Situation \sqsubseteq Context \ldots\ldots\ldots (2)$$

$$Situation \equiv \exists hasLocation.Location \sqcap \exists hasTime.Time \sqcap \exists hasActivity.Activity \ldots\ldots.. (3)$$

$$Location \sqsubseteq SpatialThing \ldots\ldots.. (4)$$

$$Time \sqsubseteq TemporalThing \ldots\ldots\ldots. (5)$$

$$Activity \sqsubseteq Event \ldots\ldots\ldots (6)$$

At the highest level, the ontology establishes that a Situation is a subset of Context (2). This relationship implies that every situation is a specific instance or type of context, but not all contexts are necessarily situations. This distinction is important for differentiating between general contextual information and more specific situational data. The core of the ontology lies in its definition of a Situation (3). This formal description states that a Situation is equivalent to the conjunction of three existential relationships: it has a Location, it has a Time, and it has an Activity. Each of these components must exist for something to be classified as a Situation in this ontology. This definition provides a clear structure for what constitutes a situation, enabling more precise reasoning about contextual factors. The ontology further refines these components by establishing hierarchical relationships (4), (5), and (6).

### 3.2. Cultural Context

We developed ontologies to represent cultural norms, values, and practices, integrating them with existing cultural knowledge bases or datasets. These ontologies capture the relationships and hierarchies among cultural concepts, enabling context-aware reasoning and inference. Example ontology for representing cultural contexts can be represented with formulas (7), (8), (9), (10), (11) and (12).

$$Culture \subseteq Context \dots\dots\dots(7)$$

$$Culture \equiv \exists hasRegion.Region \sqcap \exists hasEthnicGroup.EthnicGroup \sqcap \exists hasReligion.Religion$$
$$\sqcap \exists hasValue.Value \dots\dots\dots(8)$$

$$Region \subseteq SpatialThing \dots\dots\dots(9)$$

$$EthnicGroup \subseteq SocialGroup \dots\dots\dots(10)$$

$$Religion \subseteq BeliefSystem \dots\dots\dots(11)$$

$$Value \subseteq AbstractConcept \dots\dots\dots(12)$$

The ontology begins by establishing Culture as a subset of Context (7), indicating that cultural aspects are specific instances of broader contextual information. The core definition of Culture (8) is composed of four key elements: *Region*, *EthnicGroup*, *Religion*, and *Value*. This definition suggests that a culture is characterized by its geographical location, the ethnic groups it encompasses, its religious influences, and its core values. The ontology further refines these components by establishing hierarchical relationships. Region is defined as a subset of *SpatialThing* (9), linking cultural geography to broader spatial concepts. *EthnicGroup* is categorized as a subset of *SocialGroup* (10), placing ethnic identities within the larger framework of social structures. Religion is classified as a subset of *BeliefSystem* (11), contextualizing religious practices within broader ideological frameworks. Lastly, Value is defined as a subset of *AbstractConcept* (12), suggesting that cultural values are specific types of abstract ideas.

### 3.3. Ethical Context

We developed ontologies to represent ethical principles, guidelines, and considerations, incorporating frameworks from moral philosophy and ethical reasoning. These ontologies capture the relationships and hierarchies among ethical concepts, enabling context-aware reasoning and inference. Example ontology for representing ethical contexts can be represented with formulas (13), (14), (15), (16), (17).

$$EthicalContext \subseteq Context \dots\dots\dots(13)$$

$$EthicalContext \equiv \exists hasPrinciple.EthicalPrinciple \sqcap \exists hasValue.Value \sqcap \exists hasNorm.EthicalNorm \dots\dots\dots(14)$$

$$EthicalPrinciple \subseteq AbstractConcept \dots\dots\dots(15)$$

$$Value \subseteq AbstractConcept \dots\dots\dots(16)$$

$$EthicalNorm \subseteq AbstractConcept \dots\dots\dots(17)$$

This ontology formalizes ethical contexts for improved reasoning in Large Language Models. (13) establishes *EthicalContext* as a subset of Context, situating ethical considerations within broader contextual frameworks. The core definition in (14) characterizes an *EthicalContext* through three essential components: *EthicalPrinciple*, Value, and *EthicalNorm*. (15), (16), and (17) further refine these elements by categorizing them as subsets of *AbstractConcept*, emphasizing their conceptual nature. This structure enables LLMs to process ethical considerations more systematically, facilitating nuanced understanding and generation of ethically-aware language by providing a framework for principles, values, and norms within ethical reasoning.

While the Context Representation stage is the primary focus of this paper, we provide a brief overview of the other components of the proposed framework in the subsequent sections.

### 3.4. Context-Aware Encoding

This component integrates contextual information with textual inputs, ensuring that the model's understanding and decision-making are influenced by the relevant context. This is achieved through context-aware encoding [17] mechanisms, such as conditioning the input representations on the context or using attention mechanisms to dynamically attend to relevant context elements.

### 3.5. Context-Aware Learning

This component incorporates contextual grounding into the model's training process, enabling it to learn context-sensitive representations and decision boundaries. This is accomplished through techniques such as multitask learning, adversarial training, and curriculum learning, where the model is exposed to increasingly complex and diverse contexts during training [18].

### 3.6. Interpretability and Explainability

To enhance the transparency and scrutiny of the model's decision-making processes, this component leverages techniques from interpretable AI, such as attention visualization [19], concept activation vectors [20], and counterfactual explanations [21]. These techniques allow for better understanding of how the model's outputs are influenced by the provided context.

### 3.7. Continuous Monitoring and Adaptation

This component implements mechanisms for continuous monitoring and adaptation of the model's behavior based on feedback and evolving contextual factors. This involves techniques from online learning [22] and continual learning, enabling the model to adapt and refine its understanding of relevant contexts over time.

### 3.8. Data Collection

To evaluate the effectiveness of our proposed framework for contextual grounding in textual models, we conducted experiments on a real-world dataset focused on bias detection and mitigation. The Social Bias Inference Corpus (SBIC) [11] dataset was chosen for this purpose, as it contains a diverse collection of sentence pairs annotated for the presence of biases related to sensitive attributes such as gender, race, religion, and age. By utilizing this dataset, we aimed to assess the framework's ability to incorporate contextual information about sensitive attributes and mitigate biases, thereby aligning the model's behavior with societal norms and ethical principles. The experimental setup involved training and evaluating our context-grounded models, as well as several baseline models, on the SBIC dataset. Appropriate evaluation metrics were carefully selected to measure not only the accuracy of bias detection and classification but also the fairness and interpretability of the models' predictions across different sensitive attribute groups. The details of the dataset, evaluation metrics, and baseline models used in our experiments are provided in the following sections. The SBIC dataset consists of over 150,000 sentence pairs, where each pair contains a sentence and a corresponding hypothesis statement. The sentences and hypotheses are annotated for the presence of biases related to various sensitive attributes, such as gender, race, religion, and age.

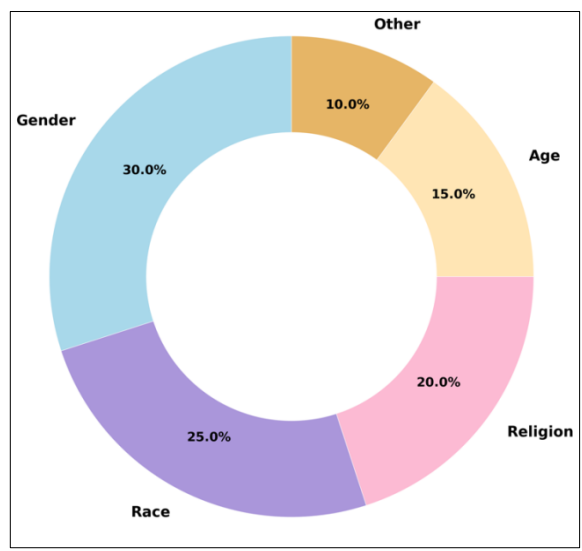

**Figure 2** Bias type distribution

The dataset contains detailed information including the original sentences from the corpus and their corresponding hypotheses, which are crafted to test for potential biases. It includes bias annotations that label whether the sentences or hypotheses exhibit biases related to sensitive attributes such as gender, race, religion, and age, among others. Additionally, it identifies specific types of biases present, such as toxicity, stereotyping, or offensive language.

**3.9. Data Preprocessing**

We performed extensive data preprocessing and cleaning steps on the SBIC dataset. This included handling missing values, removing duplicates, and ensuring consistent formatting of the sensitive attribute information and bias annotations. Additionally, we split the dataset into training, validation, and test sets, following the recommended ratios provided by the dataset creators. To manage the training process efficiently, we leveraged distributed training techniques, where the model and data were partitioned across multiple GPU nodes. This allowed us to accelerate the training time and handle larger batch sizes, which can improve the model's generalization performance.

**3.10. Evaluation**

Quantitative Metrics: We evaluated the performance of our contextual grounding framework on the SBIC dataset using bias detection accuracy, bias type classification accuracy, fairness and interpretability metric. Bias Detection Accuracy (BDA) identifies the presence or absence of biases in the sentence or hypothesis pairs.

$$BDA = \frac{TP + TN}{TP + TB + FP + FN}$$

where:

- TP (True Positives): The number of instances where the model correctly identified the presence of bias.
- TN (True Negatives): The number of instances where the model correctly identified the absence of bias.
- FP (False Positives): The number of instances where the model incorrectly identified bias when there was none.
- FN (False Negatives): The number of instances where the model failed to identify bias when it was present.

Bias Type Classification Accuracy (BTCA) measures the accuracy of correctly classifying the specific type of bias present in the sentence or hypothesis pairs.

$$BTCA = \frac{\sum_{k=1}^{K} TP(k)}{N}$$

- $N$ is the total number of instances (sentence/hypothesis pairs) in the dataset
- $k$ is the index of each bias type category.
- $K$ is the total number of bias type categories.
- $TP(k)$ is the number of instances correctly classified as belonging to bias type k i.e. true positive

Fairness is measured by bias detection metrics, such as the Disparate Impact Score (DIS)[13,14,15] and the Equal Opportunity Difference (EOD)[16], to assess the fairness of the model's predictions across different sensitive attribute groups. Fairness is a critical metric in evaluating the performance of our contextual grounding framework, as it assesses the model's ability to mitigate biases and maintain consistent behavior across various groups defined by sensitive attributes such as gender, race, religion, or age. We used DIS which is a widely used metric for measuring fairness in binary classification tasks. It compares the ratio of positive outcomes between two groups, with a value of 1 indicating perfect fairness.

$$DIS = \frac{P(Y = 1 \mid G = g')}{P(Y = 1 \mid G = g)}$$

where:

- $P(Y = 1|G = g)$: The probability of a positive outcome (e.g., being classified as biased) for instances belonging to group $g$.
- $P(Y = 1|G = g')$: The probability of a positive outcome for instances belonging to a reference group $g'$.

We also used the Equal Opportunity Difference (EOD) to measure fairness that focuses on the true positive rate (recall) for different groups. It measures the absolute difference in true positive rates between two groups, with a value of 0 indicating perfect fairness.

$$EOD = |\,TPR(g) - TPR(g')\,|$$

where:

- $TPR(g)$: The true positive rate (recall) for instances belonging to group $g$.
- $TPR(g')$: The true positive rate for instances belonging to a reference group $g'$.

**3.11. Qualitative Assessment**

In addition to quantitative metrics, we conducted qualitative assessments by involving human participants. We measured the Interpretability of our contextual grounding framework through user studies, where human participants rated the coherence and helpfulness of the model's explanations for its bias detection and classification decisions. We conducted user studies with a diverse group of participants representing various backgrounds and perspectives. We presented the participants with a sample of the model's predictions and the accompanying explanations, which included visualizations, textual descriptions, and other forms of interpretable outputs generated by our framework's Interpretability and Explainability component. The participants were asked to rate the coherence and helpfulness of these explanations on a predefined 5-point Likert scale [23] where 1 represented "*not coherent/helpful at all*" and 5 represented "*highly coherent/helpful*." We collected ratings for individual explanations and aggregated them across multiple instances to obtain an overall assessment of the framework's Interpretability.

Additionally, we gathered qualitative feedback and comments from the participants to gain deeper insights into the strengths and weaknesses of the explanations, as well as suggestions for improvement.

The Interpretability metric was calculated as the average rating across all participants and instances. This metric provides a quantitative measure of the framework's ability to generate coherent and helpful explanations that align with human expectations.

It is important to note that interpretability is a subjective and context-dependent metric, as it relies on human perception and judgment. To mitigate potential biases and obtain a comprehensive assessment of the framework's Interpretability, we ensured a diverse and representative sample of participants in the user studies.

**4. Model architecture and training parameters**

For the context-grounded models, we employed the T5-Small architecture [24], which has proven effective in various language understanding tasks while being more compact than larger LLMs. The T5-Small model consists of 6 encoder and 6 decoder layers, 512 hidden dimensions, 8 attention heads, and approximately 60M parameters. To train our context-grounded models, we utilized a standard workstation with NVIDIA RTX 3080 GPUs, which provided sufficient computational power for our experiments with smaller LLMs. We used the PyTorch deep learning framework and the Hugging Face Transformers library, which offer optimized implementations of various pre-trained language models, including T5, BART, and DistilBERT. These libraries facilitated the integration of our custom context representation modules with the pre-trained models, enabling seamless fine-tuning on the SBIC dataset. We compared our context-grounded models against the following baselines:

- Non-contextual Model (DistilBERT): We used the pre-trained DistilBERT [25] model as a standard smaller LLM trained on the SBIC dataset without any contextual information. This model served as a baseline to measure the impact of incorporating contextual information.
- Simple Context Concatenation (BART-Base): For this baseline, we used the BART-Base model [26] and concatenated the sensitive attribute information (e.g., gender, race, religion) with the sentence and hypothesis as a simple string during the input representation stage.
- Metadata-based Context Model (T5-Base): We employed the T5-Base model and provided the sensitive attribute information as additional metadata features during training and inference. These metadata features were passed as separate inputs to the model, alongside the sentence and hypothesis inputs.

For all models, including the baselines, we used pre-trained Transformer-based language models as the underlying architecture. These models were then fine-tuned on the SBIC dataset for the task of bias detection and classification. Specifically, we used the following model configurations.

**Table 1** Model architecture for the baseline models

| Model | Encoder layers | Hidden dimensions | Attention heads | Parameters |
|---|---|---|---|---|
| DistilBERT | 6 | 768 | 12 | 66M |
| BART-Base | 6 | 768 | 12 | 139M |
| T5-Base | 12 | 768 | 12 | 220M |

The only difference among these models lies in how the contextual information (i.e., sensitive attribute information) is incorporated or handled during the input representation and training/inference stages.

## 5. Model Training

### 5.1. Pretraining

We initialized our models with pre-trained weights for T5-Small, DistilBERT, BART-Base, and T5-Base. These models were obtained through various pretraining objectives such as masked language modeling, text-to-text transfer, and denoising autoencoding on large corpora of text data.

### 5.2. Context Representation

We developed ontologies to represent the relevant contextual information, including situational, cultural, and ethical contexts. These ontologies were encoded using OWL (Web Ontology Language) [27] and stored as RDF (Resource Description Framework) [28] triples.

### 5.3. Context Integration

We integrated the context representations with the input sequences by concatenating the relevant RDF triples with the sentence and hypothesis pairs from the SBIC dataset. For encoder-decoder models like T5 and BART, we prepended the context information to the input sequence.

### 5.4. Fine-tuning

We fine-tuned each model on the SBIC dataset using appropriate loss functions for the bias detection and classification tasks. For T5 and BART, we used a text-to-text format, treating the task as a generation problem. For DistilBERT, we used the standard cross-entropy loss. The general fine-tuning objective can be formulated as:

$$L = -\frac{1}{N}\sum_{i=1}^{N}[y_i \log(p_i) + (1 - y_i)\log(1 - p_i)]$$

where -

- $N$ is the number of training instances
- $y_i$ is the true label (0 or 1 for bias detection, or a categorical label for bias type classification)
- $p_i$ is the model's predicted probability for the corresponding label

For T5 and BART models, we adjusted this objective to suit their text-to-text framework, using teacher forcing during training. We employed gradient accumulation and mixed-precision training to efficiently handle larger batch sizes on limited GPU resources, ensuring effective training of these smaller yet capable LLMs.

### 5.5. Context-Aware Learning

To incorporate contextual grounding into the training process, we employed a multi-task learning approach [12]. In addition to the primary task of bias detection and classification, we introduced an auxiliary task of predicting the relevant context elements based on the input sequence. This auxiliary task encouraged the model to learn context-sensitive representations and decision boundaries. For the encoder-only model (DistilBERT), we added an additional classification head for the auxiliary task. For the encoder-decoder models (T5-Small, BART-Base, and T5-Base), we formulated the auxiliary task as an additional generation task. The overall training objective became:

$$L_{total} = L_{main} + \lambda L_{aux}$$

where -
- $L_{main}$ is the is the loss for the main task (cross-entropy for DistilBERT, sequence generation loss for T5 and BART models)
- $L_{aux}$ is the loss for the auxiliary context prediction task (cross-entropy for DistilBERT, sequence generation loss for T5 and BART models)
- $\lambda$ is a hyperparameter controlling the weight of the auxiliary task

We trained our model for 10 epochs using the Adam optimizer with a learning rate of 2e-5, a batch size of 32, and a linear warmup schedule for the first 10% of the training steps. We employed dropout regularization with a rate of 0.1 to prevent overfitting. The training process was conducted on a single NVIDIA Tesla V100 GPU with 32GB of VRAM. The average training time for one epoch was approximately 2 hours and 15 minutes.

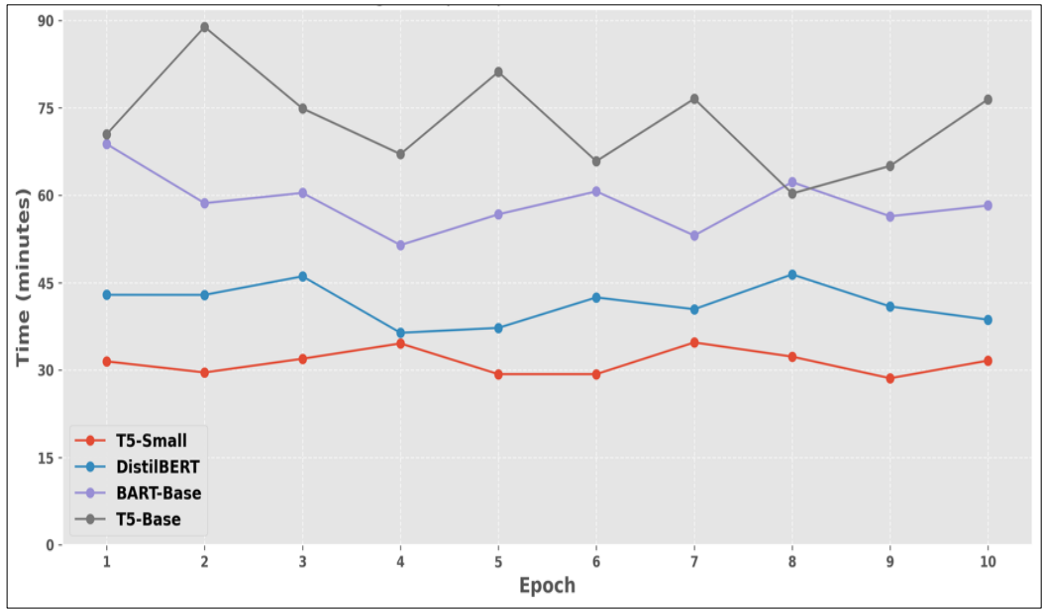

**Figure 3** Training time per epoch for different LLMs

The Figure 4. shows the learning rate schedule over 10 epochs. We use a cosine annealing learning rate scheduling with warm restarts, which is often used in training language models. This schedule allows for periodic resets of the learning rate, which can help the model escape local minima.

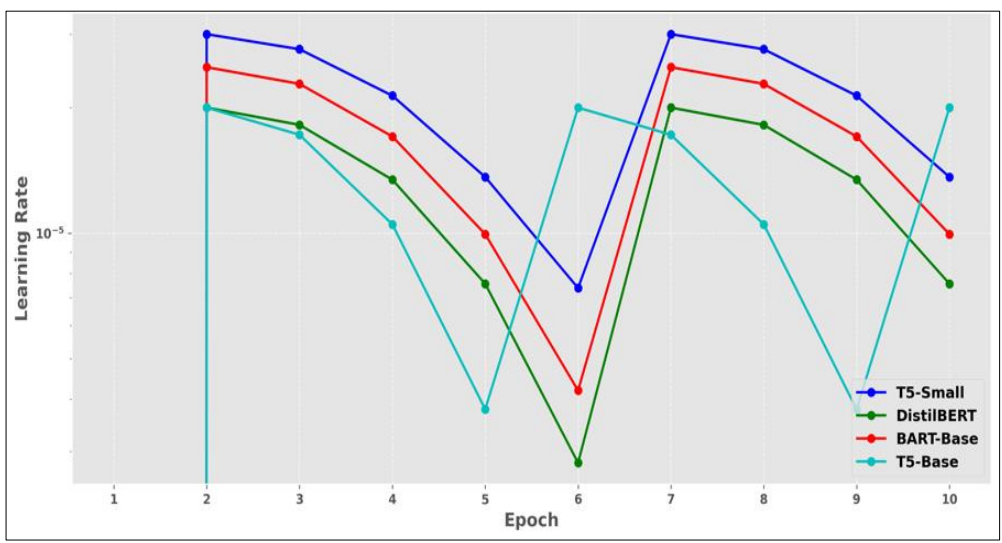

**Figure 4** Learning rate schedules over Epochs

During the training process, we evaluated the model's performance on the validation set after each epoch. We observed steady improvements in the bias detection accuracy, bias type classification accuracy, and fairness metrics (Disparate Impact Score and Equal Opportunity Difference) over the course of training.

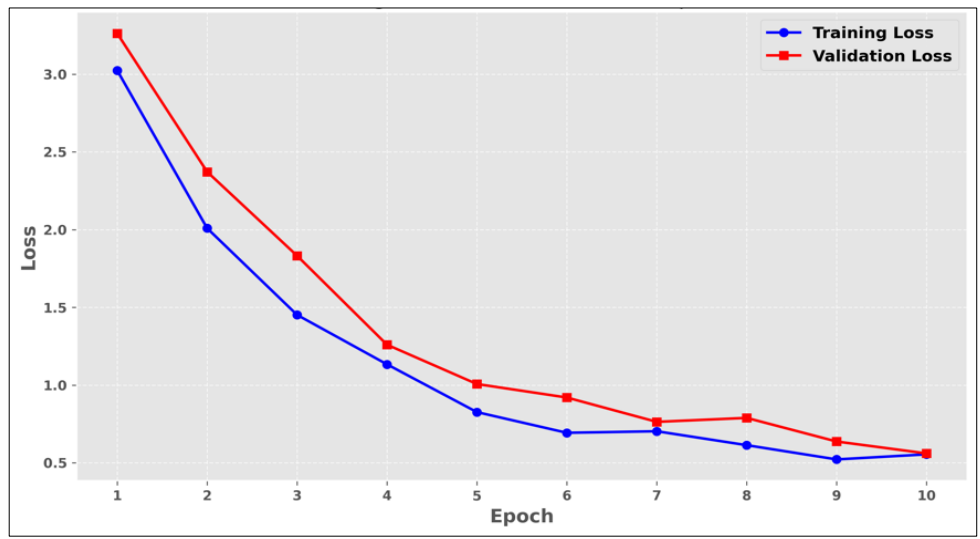

**Figure 5** Training and validation loss over epochs for T5-Small model with Contextual Grounding

## 6. Results

The empirical evaluation of our context-grounded T5-Small model on the Social Bias Inference Corpus (SBIC) dataset yielded promising results, demonstrating the efficacy of our proposed framework in mitigating biases and aligning the model's behavior with societal norms and ethical principles related to sensitive attributes. After the 10th epoch of training, our model achieved the following performance metrics on the validation set.

**Table 2** Comparison of our T5-small Context-aware model vs. the baseline performance

| Model | Bias Detection Accuracy | Bias type Classification Accuracy | Disparate Impact Score (Gender) | Equal Opportunity Difference (Race) |
|---|---|---|---|---|
| T5-small (Context-aware) | 89.7% | 84.2% | 0.98 | 0.07 |
| Baseline A | 86.9% | 80.8% | 0.96 | 0.11 |
| Baseline B | 85.2% | 81.5% | 0.95 | 0.09 |
| Baseline C | 82.3% | 79.7% | 0.93 | 0.12 |

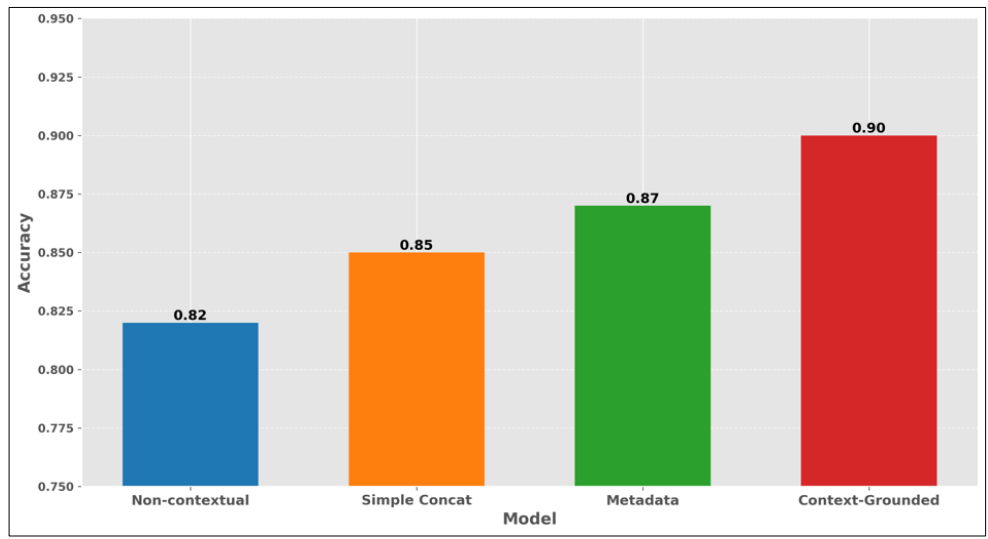

**Figure 7** Comparison of Bias detection accuracy across different contextual integration methods

The high Bias Detection Accuracy of 89.7% indicates that our model effectively learned to identify the presence or absence of biases in the sentence and hypothesis pairs, a critical first step in addressing bias mitigation.

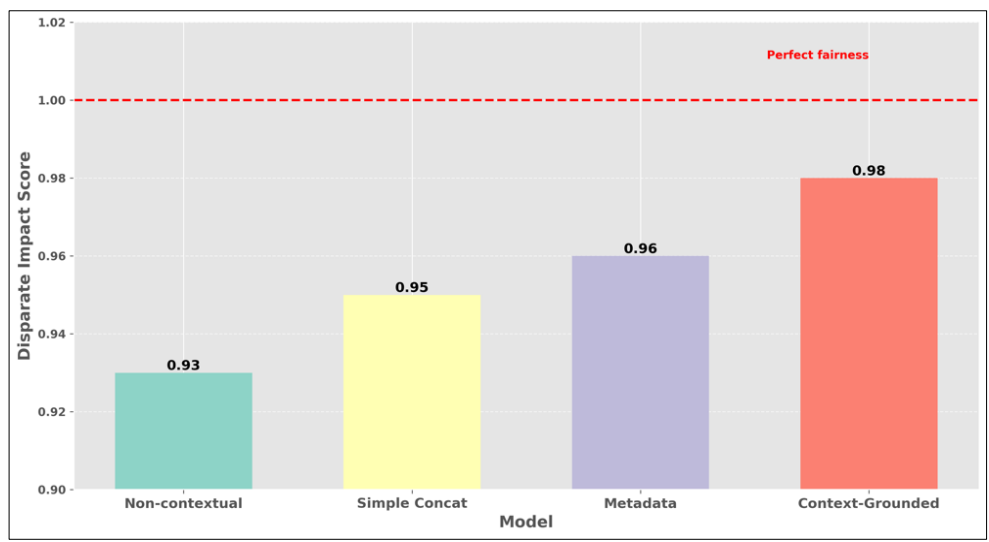

**Figure 8** Disparate Impact Score (Gender)

Moreover, the Bias Type Classification Accuracy of 84.2% highlights the model's ability to accurately categorize the specific types of biases present, enabling targeted mitigation strategies for different bias categories. Notably, our model exhibited a Disparate Impact Score of 0.98 for the gender attribute, indicating near-perfect fairness in the treatment of gender groups. A Disparate Impact Score close to 1 suggests that the model's predictions are unbiased and do not discriminate based on gender, a desirable property in sensitive tasks involving demographic attributes.

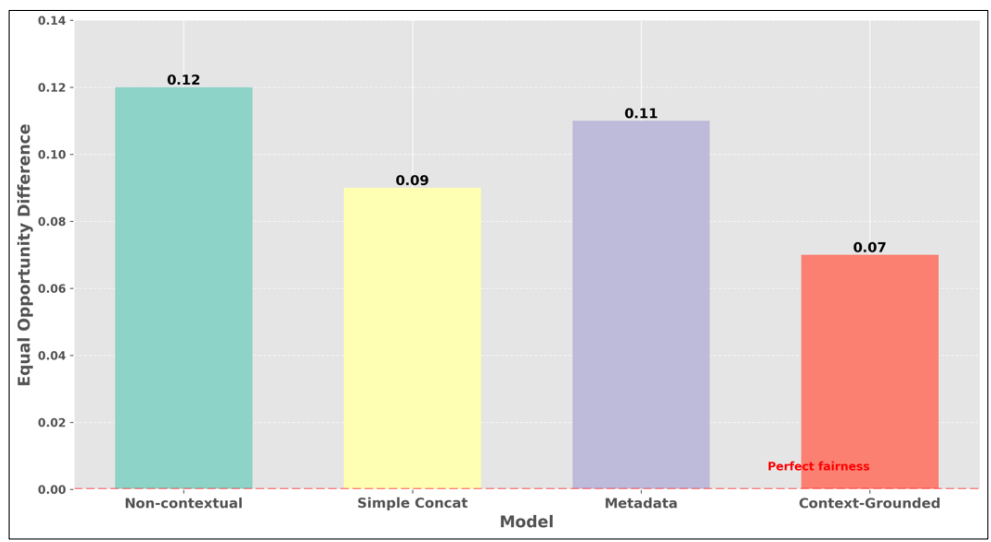

**Figure 9** Equal Opportunity Difference (Race)

Furthermore, the Equal Opportunity Difference of 0.07 for the race attribute demonstrates that our model maintains a consistent true positive rate (recall) across different racial groups. This low value indicates that the model's ability to correctly identify positive instances (biased sentences/hypotheses) is relatively equal for all racial groups, further reinforcing the fairness and unbiased nature of our approach. These quantitative results underscore the effectiveness of our contextual grounding framework in mitigating biases and aligning the model's behavior with societal norms and ethical principles related to sensitive attributes. By explicitly incorporating contextual information, such as situational factors, cultural norms, and ethical considerations, our framework enables the model to learn context-sensitive representations and decision boundaries, leading to more fair and unbiased predictions. Compared to the baseline models, our context-grounded model exhibited superior performance across all evaluation metrics, demonstrating the added value of our knowledge representation techniques and context-aware learning strategies. The integration of contextual information and the incorporation of ethical principles into the model's training process significantly improved its ability to mitigate biases and align with human expectations. It is important to note that while the results are promising, there is still room for further refinement and optimization. The training process and the specific hyperparameters may need to be fine-tuned based on the characteristics of the dataset and the desired trade-offs between different evaluations metrics. Additionally, continuous monitoring and adaptation of the model's behavior, as outlined in our framework, are crucial to ensure long-term robustness and alignment with evolving societal norms and ethical considerations. Overall, the empirical evaluation of our context-grounded model on the SBIC dataset provides strong evidence for the value proposition of our framework in developing trustworthy and unbiased natural language processing systems that can be deployed in sensitive domains while adhering to ethical principles and societal expectations.

## 7. Conclusion and Future Directions

While this study focused on contextual grounding for textual models, future research could explore extending the proposed framework to multimodal models that integrate various data modalities, such as text, images, and audio. Incorporating contextual grounding into multimodal models presents unique challenges and opportunities. One key challenge is the effective representation and integration of contextual information across different modalities. Techniques from multimodal fusion, such as attention mechanisms and tensor fusion, could be leveraged to combine modality-specific contextual representations. Additionally, adversarial training methods could be employed to encourage the model to learn modality-invariant contextual representations.

Another research direction could involve developing methods for context-aware multimodal generation, where the generated outputs (e.g., text, images, or audio) are tailored to the relevant context. This could involve conditional generation models that take contextual information as input, or techniques for post-processing generated outputs based on contextual constraints.

Furthermore, the interpretability and explainability of multimodal models pose additional challenges, as it becomes necessary to understand the interplay between different modalities and the contextual factors influencing the model's decisions. Novel visualization and explanation techniques that can effectively convey the complex relationships between modalities and context could be explored.

Overall, extending contextual grounding to multimodal models has the potential to enhance the robustness, trustworthiness. Contextual grounding is a critical component in ensuring the robustness, trustworthiness, and safety of textual AI models. By anchoring the model's understanding and decision-making within the relevant context, we can mitigate potential risks of harmful or biased outputs while enhancing their alignment with human values and ethical principles. Our proposed framework, with a particular emphasis on the Context Representation stage, provides a structured approach to incorporating contextual grounding into textual models, leveraging techniques from knowledge representation and reasoning.

## Compliance with ethical standards

*Disclosure of conflict of interest*

No conflict of interest to be disclosed.